\documentclass[12pt]{article}

\usepackage[
    backend=biber,
    style=numeric,
    sorting=none
]{biblatex}
\usepackage[hidelinks]{hyperref} 

\usepackage{graphicx}
\usepackage{amsmath}
\usepackage[normalem]{ulem}
\usepackage{hyperref}
\usepackage{times}
\usepackage{color}
\usepackage[inner=1in,
            outer=1in,
            top=1in,
            bottom=1in]{geometry}

\title{
Self-excited actuation enables adaptive and resilient flapping-wing flight
}

\author
{Rundong Yang$^{1}$, Ethan S. Wold$^{2}$, Ellen Liu$^{3}$,\\ James Lynch$^{1}$, Wei Zhou$^{1}$, Mark Jankauski$^{4}$,  Simon Sponberg$^{2,3}$, Nick Gravish$^{1}$
\\
\\
\normalsize{$^{1}$Mechanical and Aerospace Engineering,}\\
\normalsize{ University of California San Diego, San Diego, CA, 92161, USA} \\
\normalsize{$^{2}$ School of Biological Sciences and School of Physics$^{3}$}\\
\normalsize{Georgia Institute of Technology, Atlanta, GA, 30332, USA}\\
\normalsize{$^{4}$ Mechanical Industrial Engineering,}\\
\normalsize{Montana State University, MT, 59717, USA}\\
\normalsize{$^\ast$ To whom correspondence should be addressed; E-mail:  ngravish@ucsd.edu.}
}

\date{}
\begin{document}
\maketitle

\begin{abstract}
The muscles that power insect flight fall into one of two categories: 1) synchronous muscles that contract under direct control from the nervous system, and 2) asynchronous muscles which have an intrinsic stretch activation response that spontaneously generates wingbeats without the need for signaling from the brain. 
It is thought that the emergent nature of asynchronous wingbeats provides both adaptive and responsive capabilities for flight control. 
To date, most flying robots use synchronous actuation.
In this paper we develop the first flight-capable flapping wing robot that uses asynchronous actuation. 
We demonstrate that asynchronous actuation allows wings to respond to changes in the resonant mechanics of the body without control input, and wings can react instantaneously to collisions with obstacles with no extrinsic sensing needed. 
Flight tests within cluttered environments demonstrate that asynchronous actuation significantly improves stability and performance when compared to synchronous actuation.
In total this work demonstrates that a flapping wing robot actuation strategy that emulates the asynchronous muscles of flying insects can provide fast, reactive actuation responses before a control system would need to intervene. 
This partitioning of embodied control to both the low-level actuation dynamics and and high-level sensorimotor system provides a compelling blueprint for new flying robots.
\end{abstract}

\section{Introduction}

\begin{figure}[h]
\centering
\includegraphics[width=1\linewidth]{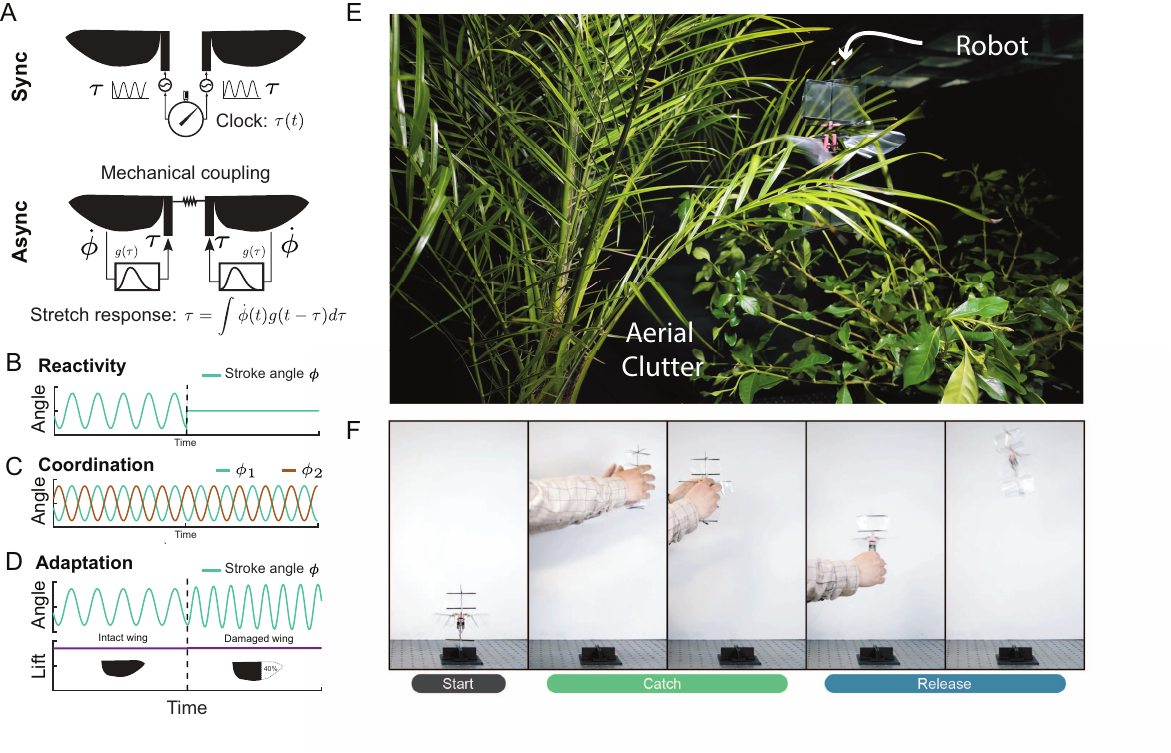}
 \caption{Overview of mechanism and capability of asynchronous actuated flapping wing robot. \textbf{A} Comparison of time-generated synchronous actuation (top) and stretch activated asynchronous actuation (bottom). \textbf{B}-\textbf{D} Three potential benefits of asynchronous actuation: \textbf{B} wingbeat collisions can inhibit ``stretch'' and halt asynchronous wingbeats quickly without contact sensors, \textbf{C} left-right wingbeat coordination through mechanical coupling can robustly synchronize motion without explicit control, and \textbf{D} asynchronous actuation provides wingbeat frequency adaptation in response to wing-damage. \textbf{E} Photo of our robot in a naturalistic cluttered aerial environment stably hovering amidst continuous wing collisions. \textbf{F} Time series of catching a free-flying robot which automatically halts wingbeats and releasing back into flight.}
\label{fig:intro}
\end{figure}

Flying insects are capable of some of the most remarkable feats of locomotion in the natural world. 
To achieve these capabilities, many insects use large power muscles to drive rapid wingbeats and small steering muscles to modulate wing orientation \cite{Dickinson1997-nx, Dudley2002-kl}. 
Furthermore, insects are divided by the way their power muscles are actuated (Fig.~\ref{fig:intro}A). 
Insects with synchronous muscle rely on a neural signal to initiate each contraction, and thus the wingbeat is directly timed by the brain.
In contrast, insects with asynchronous muscle generate self-excited oscillations without the need for neural input, instead a stretch activation property intrinsic to the muscle results in emergent oscillations \cite{josephson2000asynchronous, pringle1949excitation}. 
A plethora of common flying insects including flies, bees, and mosquitoes, have asynchronous muscle that produces wingbeat actuation. 
To date, all flight-capable flapping-wing robots use synchronous actuation from a flight controller to generate wingbeats \cite{Agrawal2026-kc}. 
It remains an open question what potential opportunities flapping-wing robots might gain by instead employing an asynchronous actuation scheme such as exhibited by some of the most agile insects.

Synchronous actuation offers the benefit of direct control over wingbeat kinematics \cite{fei2019learning, steinmeyer2019yaw}. 
This approach to wingbeat control has led to remarkable capabilities in flapping wing robots including autonomous flight through clutter \cite{Chen2021-ui}, complex aerial maneuvers  \cite{Hsiao2025-lz}, landing on vertical and inverted surfaces  \cite{Hyun2025-gc, Chirarattananon2014-yo}, and even multimodal flight such as submerging in and emerging out of water  \cite{Zufferey2026-lu, Chen2017-jw}. 
For example, to maintain efficient wing motion as body and wing resonance properties change, synchronously actuated systems typically rely on continuous sensing of wingbeat kinematics and feedback control \cite{zhang2017resonance,Jafferis2016-le,Zhang2013-ex}. Similarly, if wings collide with an object, a synchronous actuation scheme needs to detect this collision and halt wingbeat generation. This is in contrast to asynchronous actuation, which establishes wingbeats as a limit-cycle process that is dependent only on the current state of motion of the wings and not an external frequency generator \cite{Gau2023-qs, Lynch_undated-fy}. Limit-cycles are a hallmark of nonlinear dynamical systems and they possess novel adaptive properties that have been well characterized and utilized in robotics \cite{Ajallooeian2013-oh, Buchli2006-nr, Sproewitz2008-vn}. 

Asynchronous actuation presents several opportunities for flapping wing robots such as enabling long-term resonance tracking \cite{Wold2025SupraResonant}, state-dependent actuation that instantaneously responds to changes in the internal or external state of the system \cite{mingjing2021asynchronous}, and reducing computational overhead as wingbeat generation is offloaded to the actuator itself rather than a higher-level control system. In insects, asynchronous actuation arises not from a centralized controller but from the delayed stretch activation property of the muscle itself \cite{josephson2000asynchronous, pringle1949excitation}. 
When the power muscle is stretched, a contractile force is developed that causes the muscle to shorten. 
This force is developed over a delayed time in response to the stretch and is thus called delayed stretch activation (dSA).
Two muscles arranged antagonistically thus generate oscillations: as the downstroke muscle shortens, it stretches the upstroke muscle which causes it to develop force and shorten, which stretches the downstroke muscle and the cycle continues indefinitely. 
Realizing delayed stretch activation in a flight-capable robot requires solving numerous technical challenges to design and implement delayed stretch activation in an engineered system (Fig.~\ref{fig:intro}B-D). 
However, actuation that emulates the asynchronous power muscles of insects may enable improved flight in aerial clutter and responsive wing motion through collisions and contact (Fig.~\ref{fig:intro}E-F).

In this paper, we present the first flight-capable flapping wing robot that uses asynchronous actuation to drive emergent wingbeats. 
We address this challenge by developing an actuation method to emulate delayed stretch activation using electromagnetic motors and proprioceptive stretch sensing, allowing the actuator to sense its own motion and generate delayed stretch activation without additional dedicated sensors. 
We demonstrate that this actuation strategy allows the wings to passively track changes in the resonant mechanics of the body without control input, and to react instantaneously to collisions with obstacles without extrinsic sensing. 
We further show that a simple bioinspired method of elastic mechanical coupling enables synchronization between the left and right wings rather than a feedback control scheme. 
Lastly, we demonstrate the first free-flight of an asynchronous flapping wing robot and demonstrate performance advantages within cluttered environments that significantly improve stability and flight performance compared to conventional synchronous actuation.

\newpage
\section{Results}

\subsection*{Robot design} 

\begin{figure}[h]
\centering 
\includegraphics[width=1\linewidth]{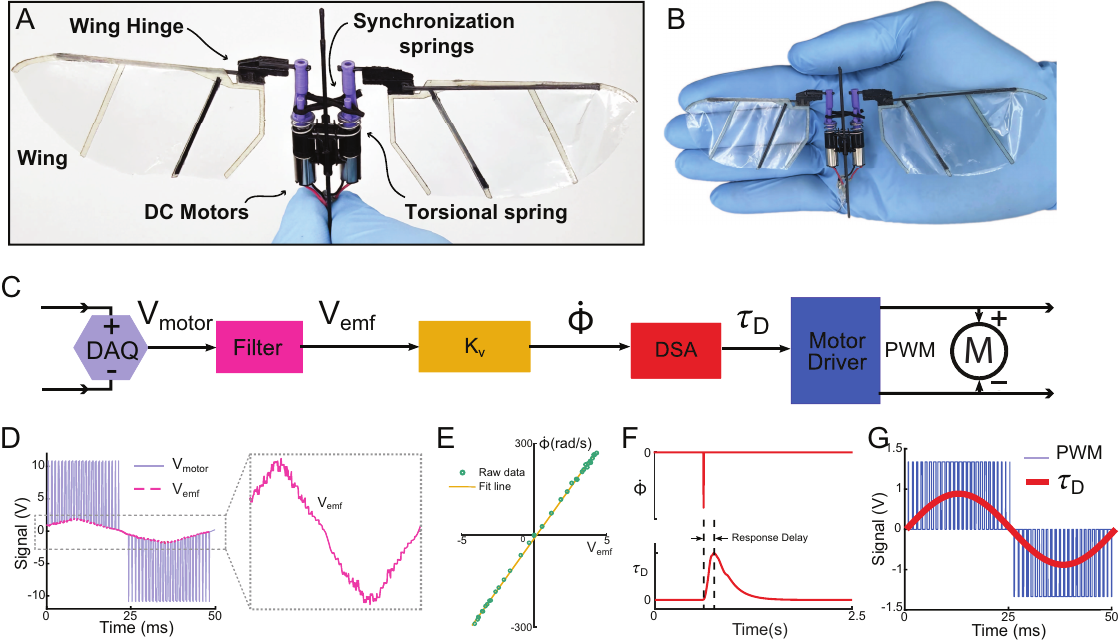}
 \caption{Robot design and implementation of sensorless asynchronous actuation in a DC motor. 
 \textbf{A-B} Robot components and scale. 
 \textbf{C} Closed-loop feedback process flow for asynchronous actuation.
 \textbf{D} We first measure motor terminal voltage and extract the back EMF voltage ($V_{EMF}$).
\textbf{E} Back EMF voltage is converted to angular speed through a calibration measurement curve.
\textbf{F} The delayed stretch activation torque ($\tau_D$) is computed through the DSA impulse response function with angular speed as input. 
\textbf{G} DSA torque is converted to PWM duty factor by the motor driver and sent to the DC motor. 
The motor terminal voltage is measured by the DAQ in a closed-loop feedback. 
 }
\label{fig:dsaimplement}
\end{figure}

We first developed a 5~g flapping wing robot to implement asynchronous actuation (Fig.~\ref{fig:dsaimplement}A, B). 
The robot has a wing span of 155~mm and the wings are constructed from a reinforced carbon fiber frame with a thin mylar membrane. 
The wings are able to passively pitch via a hinged joint that limits pitching to $\pm45^\circ$. 
Two DC motors (Pololu 26:1 Sub-Micro Plastic Planetary Gearmotor; 1.25~g, 6~mm diameter) were used to independently actuate the left and right wings.
This actuation scheme mimics the separate left and right wing power muscles in asynchronous insects \cite{Dickinson1997-nx}. 
The motors were controlled by pulse-width modulation (PWM) signals generated by two motor drivers (Pololu TB9051FTG Single Brushed DC Motor Driver Carrier) supplied at 15~V. 
Under these operating conditions, the robot generated a maximum lift force of 91.2~mN (1.8 times body-weight; Fig.~S1).
The robot is controlled through an electrical tether (4x Remington 39 AWG wires) which transmits PWM signals and enables back-electromotive force (back-EMF) measurement.
A steel coil spring is connected between the motor shaft and the robot body and thus acts as an elastic element in parallel with each the wing.

To synchronize wing motion across the two independent left and right wings, we emulated the mechanical synchronization method demonstrated in flies: an elastic structure on the thorax between left and right wings couples wingbeat phases \cite{Deora2021-hy,Deora2015-ii}.
In our robot the two motors are mechanically coupled with springs instead of through electrical control. 

To aid passive flight stability, the aerodynamic center of pressure for each wing is positioned above the robot's center of mass. 
In addition, two air dampers are mounted symmetrically above and below the chassis, resulting in a total robot height of 200~mm. 
These dampers generate stabilizing aerodynamic drag forces when the robot is perturbed from the upright orientation, providing open-loop attitude stability during free flight \cite{Fuller2017-cr}.
This mechanical stabilization scheme enabled us to focus on the dynamics of asynchronous actuation, similar to how insects separate their large asynchronous wingbeat power muscles from the smaller directly controlled steering muscles \cite{Dickinson1997-nx}.

\subsection*{Sensorless velocity measurement through motor back‑EMF} 

To implement delayed stretch activation in a flying robot, we use a Simulink Desktop real-time impulse response block that takes in motor speed and outputs a delayed stretch activated motor voltage (Fig.~\ref{fig:dsaimplement}C).
Thus, we require real-time measurement of each wing's angular velocity. 
Instead of adding extra sensors such as an encoder to the robot, we co-opted the self-generated back-EMF voltage $V_{\text{emf}}$ of the DC motors which is proportional to rotational velocity (Fig.~\ref{fig:dsaimplement}E). 
During flight, the voltage across the motor terminals is either a high-voltage PWM pulse or the velocity-dependent back-EMF signals (Fig.~\ref{fig:dsaimplement}D). 

To simultaneously drive the motor and measure back-EMF through the same terminals, we employ a motor driver capable of switching between a PWM drive mode and a high-impedance measurement mode. 
In measurement mode, the motor terminals are electrically isolated from the drive voltage, allowing the back-EMF to be measured across the motor terminals.
To extract just the back-EMF voltage we use an edge detection scheme in which we measure the voltage between the PWM off and PWM on edges. 
The back-EMF voltage is interpolated between measurements and then  converted to rotational speed through the motor constant. 
Since speed measurement requires the PWM signal to be off, we enforced an upper limit of the allowable PWM duty factor to 0.7 (See Materials and Methods).

\subsection*{Asynchronous actuation principle}

When an asynchronous muscle is subject to a sudden increase in length (an impulse in speed), the actuator gradually develops a contractile force that acts against the direction of stretch (Fig.~\ref{fig:dsaimplement}F). 
Under a repeating sinusoidal strain, the delayed stretch activation force is a phase-delayed sinusoid that can generate positive power in appropriate parameter regimes. 
Thus, the combination of elastic body elements and actuator are sufficient to produce self-sustained asynchronous oscillations (Fig.~\ref{fig:dsaimplement}G) which emerge through a hopf bifurcation in the system parameters \cite{Gau2023-qs, Lynch_undated-fy}.

To understand the operational parameter range for robotic asynchronous flight, we model robot and actuation dynamics as a coupled system.
The wingstroke ($\phi$) dynamics are generated by the motor torque and balanced by the wing's inertial force, the elastic restoring force from the spring, and the aerodynamic force as given below \cite{Gau2022-ph, Gau2023-qs}
\begin{align}
    I \ddot{\phi}+\Gamma|\dot{\phi}| \dot{\phi}+k_\phi \phi= \mu \tau_D
\label{eqn:wingdynamic}
\end{align}
The model parameters are the actuation gain $\mu$, dSA torque $\tau_D$, lumped wing and motor inertia $I$, time-averaged aerodynamic drag coefficient $\Gamma$, and spring constant $k_\phi$. 
To simplify the analysis, the model does not explicitly capture the wing-pitching dynamics. 
The full dynamics of the passively pitching wing are incorporated in Supplementary S1 to determine the required pitching stiffness.

Motivated by asynchronous muscle studies \cite{pringle1949excitation, josephson2000asynchronous, Lynch2022-yf}, we model dSA as a second-order linear system parameterized by rate constants $r_3$ and $r_4$ which represent the rate of tension increase and decrease in response to a stretch. 
We define the variable $\kappa = r_4 / r_3$ to represent the ratio of tension decay to onset rates. 
The asynchronous actuation torque $\tau_D$ evolves according to
\begin{align}
    \ddot{\tau}_D + r_3 (1 + \kappa) \dot{\tau}_D + \kappa r_3^2 \tau_D = - \kappa r_3^2 \dot{\phi},
\label{eqn:dsa}
\end{align}
Equations~\ref{eqn:wingdynamic} and~\ref{eqn:dsa} form a coupled dynamical system which is capable of generating limit-cycle oscillations. 
Linear stability analysis shows that oscillations emerge whenever the dSA gain, $\mu$, exceeds the critical threshold.
\begin{align}
\mu > Ir_3(1 + \kappa)\left(1 - \frac{k_{\phi}/I}{\kappa r_3^2}\right)
\label{eq:instab}
\end{align}
This stability analysis provides a general criterion for the emergence of oscillations in systems governed by Eqs.~\ref{eqn:wingdynamic} and~\ref{eqn:dsa}.
\noindent Since $\mu$, $\kappa$, $I$, and $r_3$ all are strictly positive values, the relationship between system natural frequency $\omega_n = \sqrt{k_{\phi}/I}$ and dSA timescale $r_3$ is bounded between $0 < \frac{k_{\phi}/I}{\kappa r_3^2} < 1$.
In essence, the dSA response rate must always be larger than the mechanical timescale for limit-cycle asynchronous wingbeats.

\subsection*{Asynchronous actuation is self‑excited and adaptive} 
\begin{figure}[h]
\centering
\includegraphics[width=1\linewidth]{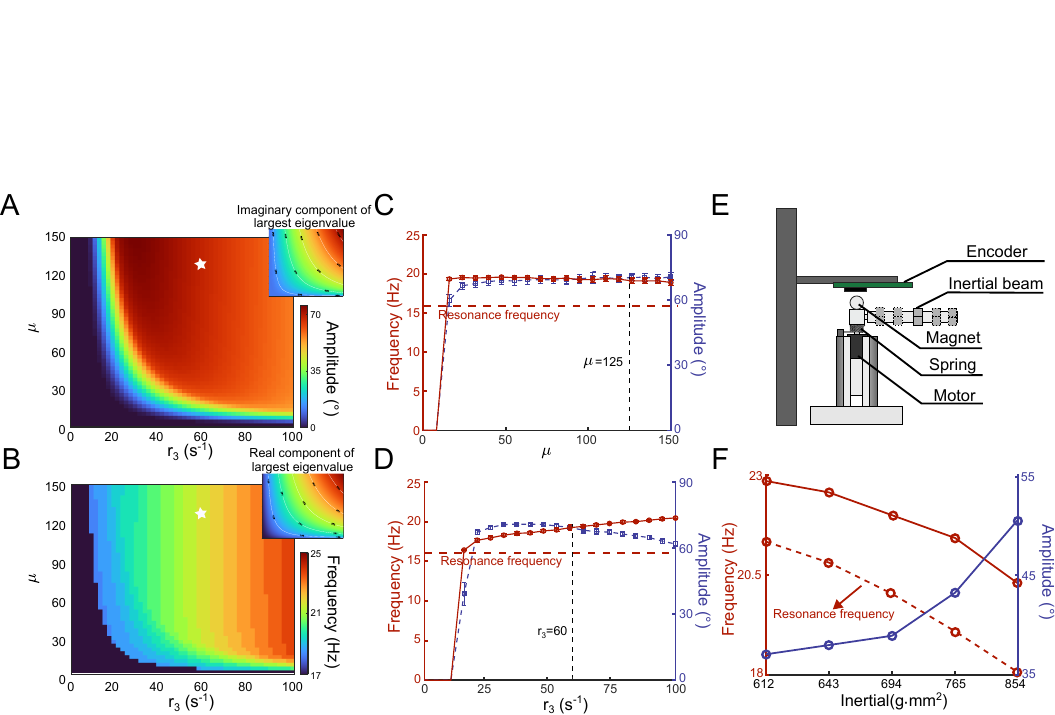}
 \caption{Effects of dSA parameters and inertia on emergent asynchronous wingbeat. \textbf{A} \textbf{B} Simulated emergent wingbeat with different combination of $\mu$ and $r_3$. Insets are the eigenvalues of the linearized robot system with asynchronous actuation. The star marker shows the values we used for our robot. \textbf{A} Wingbeat amplitude \textbf{B} Wingbeat frequency \textbf{C} Influence of $\mu$ on experimental wingbeat frequency and amplitude when $r_3=60 s^{-1}$. \textbf{D} Influence of $r_3$ on experimental wingbeat frequency and amplitude when $\mu=125$.  \textbf{E} A diagram showing the experiment setup for inertial sweep test. \textbf{F} The influence of inertia on the wingbeat with asynchronous actuation. Dashed line represents the resonance frequency under certain value of inertial.}
\label{fig:eigenvalues}
\end{figure}
To quantify how dSA parameters influence oscillation behavior, we first characterize the resonance properties of the spring-wing system. 
The reflected inertia of the DC motor with gearbox is experimentally measured to be $I_m = 541~\mathrm{g\cdot mm^2}$ (Fig.~S2), and the torsional spring constant $k_\phi = 12.9~\mathrm{N\cdot mm/deg}$. 
These values predict an undamped natural resonance frequency of $24.1~\mathrm{Hz}$ in the absence of wing aerodynamics. 
We also experimentally measured the stroke displacement resonance frequency with an attached wing to be $15.5 ~\mathrm{Hz}$ ($n = 5$) (Fig.~S3) indicating that the presence of aerodynamic damping lowers the resonance frequency \cite{martinez2023fluid, Lynch2021-fn}.

We then performed a parameter sweep of Eqs.~\ref{eqn:wingdynamic} and \ref{eqn:dsa} using the experimentally measured parameters. 
The delayed stretch activation (dSA) gain ($\mu$) and stretch activation rate ($r_3$) were varied systematically, and the resulting steady-state oscillation frequency and amplitude were computed (Fig.~\ref{fig:eigenvalues}A,B). 
Because the asynchronous dSA response was normalized following Gau \textit{et al.}~\cite{Gau2023-qs}, the simulations isolate the effects of response timing and delay on flight dynamics without introducing changes in the overall force magnitude. 
Consequently, the parameter-sweep results differ slightly from those reported previously~\cite{Lynch2022-yf} regarding the overall trend of the wingbeat across different dSA parameters.   

In our system, oscillations were not sustained when either parameter approached zero, owing to physical dissipative effects including friction, aerodynamic drag, and finite motor torque, as well as failure to satisfy the instability criterion (Eq.~\ref{eq:instab}). 
Within the oscillatory regime, $r_3$ strongly influenced both the emergent frequency and amplitude. 
As $r_3$ increased, the oscillation frequency rose monotonically from approximately $17~\mathrm{Hz}$, slightly above the system's stroke displacement resonance frequency Fig.~\ref{fig:eigenvalues}D. 
In contrast, the oscillation amplitude increased rapidly near the onset of instability and then continued to grow more gradually at larger $r_3$ values.

Theoretically, $\mu$ scales the force output and thus the wingbeat amplitude; however, its practical effect is limited by implementation constraints. 
In particular, increasing $\mu$ ceases to affect wing kinematics once the commanded torque reaches the duty-cycle saturation limit. 
To achieve large-amplitude, high-frequency flapping, we selected operating values of $\mu = 125$ and $r_3 = 60~\mathrm{s^{-1}}$ within the feasible regime highlighted in Fig.~\ref{fig:eigenvalues}A, B. 
These operating parameters yield a nominal wingbeat frequency of $22~\mathrm{Hz}$ with a stroke amplitude of $68^\circ$ in simulation. 
We experimentally studied the sensitivity of frequency and amplitude to dSA parameters by fixing $r_3 = 60~\mathrm{s^{-1}}$ while sweeping $\mu$, and by fixing $\mu = 125$ while sweeping $r_3$, as shown in Fig.~\ref{fig:eigenvalues}C, D. 
The experimentally observed trends in both frequency and amplitude closely match those predicted by the simulation heatmaps. 
At larger parameter values, the measured frequency saturates at approximately $19.5~\mathrm{Hz}$ due to physical constraints such as motor torque limits and aerodynamic loading, supporting the results and validity of the model.

A potential advantage of asynchronous actuation is that the emergent wingbeat is set by the dSA and spring-wing parameters and not externally timed. 
We performed an experimental test to observe adaptive wingbeat frequencies as the system's mechanical resonance frequency changes.
We use an inertial beam (Fig.~\ref{fig:eigenvalues}E) in place of a wing to eliminate aerodynamic effects and then varied the beam inertia $I_{\text{load}}$ from $612~\mathrm{g\cdot mm^2}$ to $854~\mathrm{g\cdot mm^2}$, resulting in a change in resonance frequency from $23~\mathrm{Hz}$ to $20.5~\mathrm{Hz}$. 
The experimental results (Fig.~\ref{fig:eigenvalues}F) show that the emergent asynchronous wingbeat frequency tracked the changing mechanical resonance frequency (yet always remained slightly above resonance).
Furthermore, the flapping amplitude increased with inertia. 
Notably, similar results have been observed in asynchronous insects: honeybees with damaged wings increase flapping frequency in proportion to the inverse square of changes in wing inertia, consistent with tracking mechanical resonance \cite{greenewalt1960wings}. 
These results demonstrate that dSA provides an adaptive actuation mechanism that can respond to changes in the robot's mechanical resonance. 

\subsection*{Single‑wing collision attenuation with asynchronous actuation} 

Since asynchronous actuation is reliant on wing motion, when a wing collides with an obstacle and stops, the subsquent actuation may be stopped as well.
To investigate this behavior, we conducted single-wing collision experiments across synchronous and asynchronous actuation modes. 
The robot starts at a steady-state flapping motion and then a rigid obstacle is introduced at prescribed angular positions along the wing stroke (Fig.~S4).
The obstacle is instrumented with a force sensor (Sparkfun 100g Load Cell, TAL221) to measure collision force and the signal was low-pass filtered using a Kaiser-window FIR filter with a 24 Hz passband edge and 60 dB stopband attenuation.

\begin{figure}[h]
\centering
\includegraphics[width=1\linewidth]{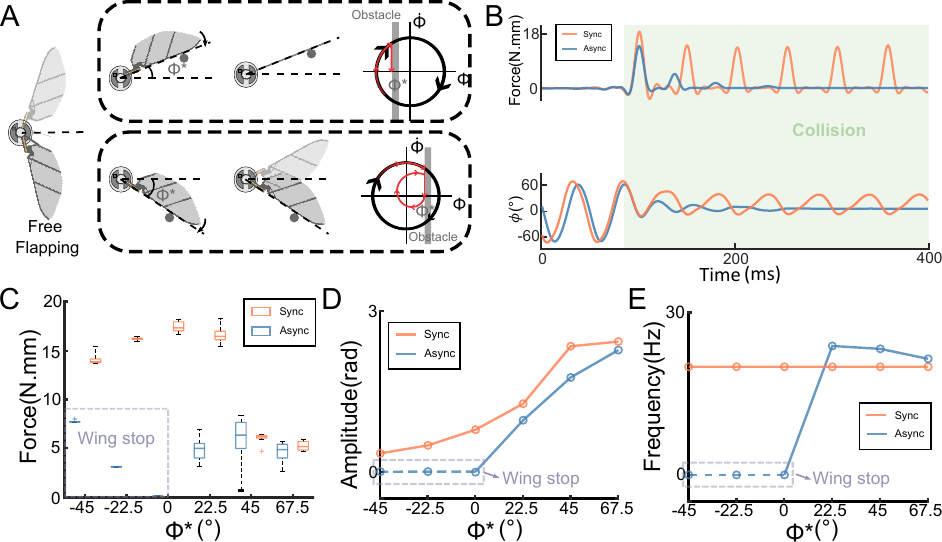}
\caption{Comparison of collision responses in asynchronous and synchronous actuation. \textbf{A} A diagram showing the response of wingbeat with dSA at different collision angles. To the right are trajectories of the wingbeat in phase space, where black lines represent the ideal limit cycle generated by dSA without perturbation. Red lines represent the actual trajectory when collision happens. \textbf{B} Collision force and wing angle of asynchronous-actuated and synchronous-actuated wingbeat VS time when the collision happens at the neutral position. Green region represents when collision starts and remains. \textbf{C}-\textbf{E} Comparison of averaged collision force, amplitude and frequency during collision at steady state between asynchronous-actuated and synchronous-actuated wingbeat at different collision angles respectively.}
\label{fig:collision}
\end{figure}

Fig.~\ref{fig:collision}B shows a representative case where the collision occurs at the neutral position ($\phi^* = 0$). 
Upon impact, the wing driven by asynchronous actuation rapidly stops, as evidenced by an abrupt reduction in both angular amplitude and collision force to near zero. In contrast, the synchronously actuated wing, driven by a sinusoidal command, \(0.7\sin(\omega t)\), continues flapping because the prescribed time-dependent control signal persists after impact. Consequently, it generates substantially larger collision forces, with subsequent impacts reaching approximately (80\%) of the initial impact force.

The collision-response depends strongly on the wingbeat phase at which contact occurs (Fig.~\ref{fig:collision}A). 
Figures~\ref{fig:collision}C--E summarize the averaged collision force, flapping amplitude, and flapping frequency during collision at steady state as functions of collision position $\phi^*$. 
When collisions occur during the late downstroke (\(\phi^* \geq 45^\circ\)), the collision forces are comparable for synchronous and asynchronous actuation. Particularly, as the collision point approaches stroke reversal, where the wing velocity is low, the impact forces are small and with little difference between synchronous and asynchronous actuation. 
In contrast, for collisions occurring near or before mid-stroke (\(\phi^* \leq 22.5^\circ \)), asynchronous actuation reduces the collision force by more than threefold compared with synchronous actuation.
Furthermore, once the collision angle exceeds the neutral wing position, the asynchronously actuated wing comes to a complete stop, whereas the synchronously actuated wing continues pushing against the obstacle. 

In the regime where the asynchronously actuated wing continues flapping (\(\phi^* \geq 22.5^\circ\)), the flapping frequency increases slightly from \(21.4\,\mathrm{Hz}\) to \(23.5\,\mathrm{Hz}\) with the decrease of collision angle \(\phi^*\). 
This increase arises because the collision-induced reversal of wing velocity immediately changes the sign of the actuator torque, allowing the actuator dynamics to respond directly to the perturbation. 
In comparison, synchronous actuation continues to follow the prescribed time-based input and is therefore insensitive to the instantaneous wing state. The flapping amplitude decreases with increasing collision phase for both actuation modes as geometric constraints progressively limit wing motion. However, the reduction is less pronounced under asynchronous actuation because the actuator torque remains coupled to the wing dynamics and can adapt to external perturbations.

In summary, under synchronous actuation, the wing exhibits minimal velocity disturbance during collisions and continues to generate the prescribed driving force. 
In contrast, asynchronous actuation consistently suppresses speed overshoot and post-impact oscillations across repeated collisions. 
Together, these experiments (this and last section) demonstrate that dSA enables wingstrokes to adapt automatically to both changes in mechanical resonance and impulsive disturbances such as collisions. 
These properties are highly advantageous for flight in complex and uncertain environments. 

Notably, similar behaviors have been observed in insects employing asynchronous muscle. 
Wing collisions in insect flight can cause abrupt halting and re-initiation of wingbeats that closely resemble the response observed in our Robophysical system \cite{Lynch2022-yf}. 
These parallels suggest that dSA can act as control response through its inherent material response—or, in our case, actuator dynamics—without the the need for centralized control or additional sensing.

\subsection*{Mechanical synchronization of independently actuated wings} 
 \begin{figure}[]
\centering
\includegraphics[width=1\linewidth]{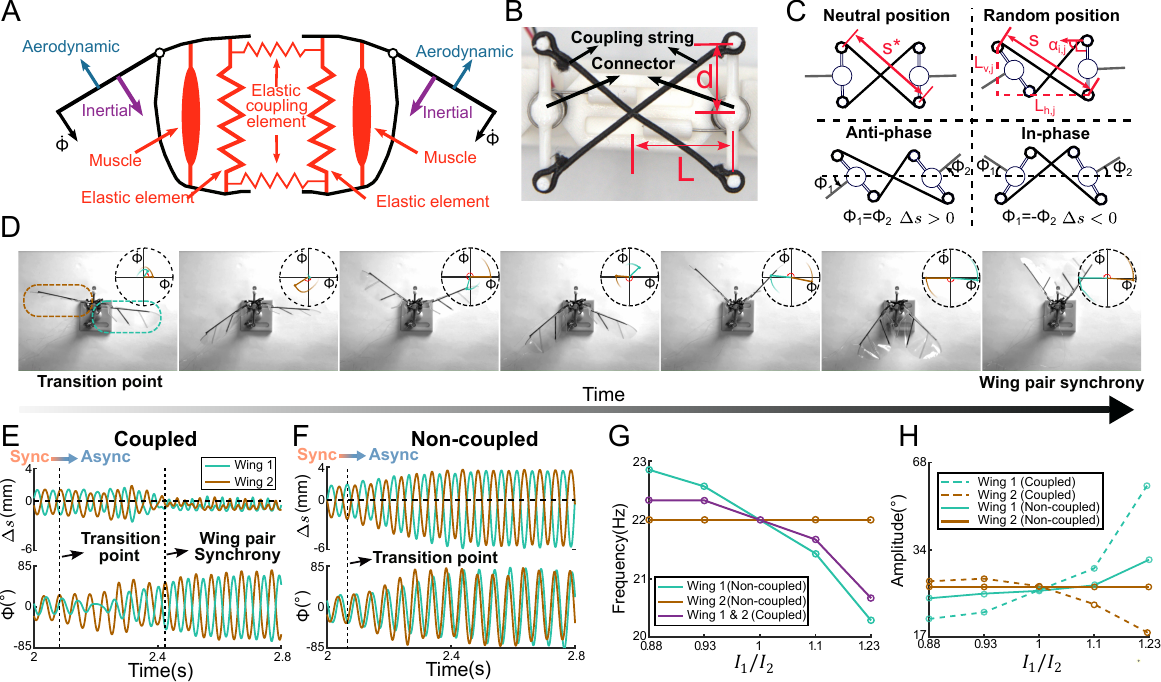}
\caption{Properties of bioinspired elastic coupling for emergent asynchronous wingbeat. \textbf{A} Schematic illustration of the elastic coupling mechanism found in the insect thorax (Image inspired from the fly thorax \cite{Deora2015-ii}). 
\textbf{B} Photograph of a benchtop version of the coupling mechanism showing the geometric parameters \(L\), \(d\). 
Two elastic coupling strings connect opposite diagonal points of the connectors. \textbf{C} Schematic of the coupling mechanism showing the geometric parameters \(s^*\), \(s\), \(L_{v,j}\), and \(L_{h,j}\), as well as the configurations of the coupling strings under in-phase and anti-phase motion. 
\textbf{D} Snapshots showing the transition from synchronous to asynchronous actuation in the coupled robot. 
During the transition, the phase difference converges from \(\pi/3\), the prescribed value during synchronous actuation, to \(\pi\). 
\textbf{E, F} Time histories of coupling-string displacement \(\Delta s\) and wing angle \(\phi\) during the transition from synchronous to asynchronous actuation in the coupled (\textbf{E}) and non-coupled robots (\textbf{F}). \textbf{G, H} Comparison of coupled and uncoupled robots under asymmetric wing inertia. 
(\textbf{G}) Flapping frequency as a function of inertia ratio. 
(\textbf{H}) Flapping amplitude as a function of inertia ratio.}

\label{fig:coupling}
\end{figure} 

Stable flight requires that wings synchronize their flapping frequency and phase while still retaining the ability to independently adapt and respond to collisions or changes in the mechanics of an individual wing. 
To achieve synchronization between two independent actuators without additional sensing or feedback control delays, we adopted a bioinspired strategy in which the left and right wings are elastically coupled, analogous to the elastic thoracic coupling mechanism found in flies (Fig.~\ref{fig:coupling}A) \cite{Deora2015-ii}. Similarly, our robot incorporates elastic interconnections between the wings to couple their flapping dynamics, enabling passive synchronization while maintaining adaptability to asymmetric perturbations \cite{Deora2021-hy, Deora2015-ii, Hao2025-xq}.

The mechanical synchronization mechanism is illustrated in Fig.~\ref{fig:coupling}B. 
It consists of two elastic springs that cross-connect the left and right actuators. 
Each motor shaft is attached to a lever arm of length $d$, with spacing between each motor of length $2L$ (Fig.~\ref{fig:coupling}B). 
As the wings flap, phase differences between the two actuators cause the coupling springs to stretch, generating a restoring torque on each wing that promotes synchronization.

The neutral length of each coupling spring is defined as:

\begin{equation*}
s^* = \sqrt{(2d)^2 + (2L)^2}.
\end{equation*}

During flapping, the extension of coupling spring $j$ is determined by the instantaneous distance between its two attachment points:

\begin{equation*}
\Delta s_j = \sqrt{L_{h,j}^2 + L_{v,j}^2} - s^*,
\end{equation*}

\noindent where $L_{h,j}$ and $L_{v,j}$ represent the instantaneous horizontal and vertical distances between the two connecting points of the coupling spring $j$, respectively. These quantities are given by

\begin{align*}
L_{v,1}=L_{v,2} &= d(\cos\phi_1+\cos\phi_2), \\
L_{h,1} &= d(-\sin\phi_1-\sin\phi_2)+2L, \\
L_{h,2} &= d(\sin\phi_1+\sin\phi_2)+2L.
\end{align*}

The corresponding coupling torque applied to wing $i$ by spring $j$ can then be expressed as

\begin{equation}
T_{i,j}(\phi_1,\phi_2)
= dk_c\cos\alpha_{i,j}\,\Delta s_j,
\label{eqn:coupling}
\end{equation}

\noindent where $k_c$ is the spring stiffness and $\alpha_{i,j}$ is the angle between the spring force direction and the direction perpendicular to connector arm $i$. 
To achieve robust in-phase synchronization while accommodating parameter mismatches between the two wings and satisfying the geometric constraints of the robot (Fig.~S5), the coupling mechanism was designed according to the parameters listed in Table~S1. 
Linearized stability analysis (Supplementary Text~S4) shows that the resulting system consistently converges to a stable in-phase motion over a broad range of wing parameter variations.

For normal flapping wing motion \( \phi_1 = -\phi_2 \), the coupling springs remain slack (\( \Delta s_j \leq 0 \)) and therefore do not exert any force (Fig.~\ref{fig:coupling}C). 
For the geometry of our flapping this is a stable state (Supplementary Text~S4), and the two wings will oscillate together at the coupled spring-wing resonance. 
When the wings become anti-phase (\( \Delta s_j > 0 \)), the coupling springs stretch and generate restoring torques. 
These torques interact with the adaptive limit-cycle dynamics produced by asynchronous actuation, driving the system back toward symmetric flapping. 
Experimental results in Fig.~\ref{fig:coupling}D-F show the transition from synchronous actuation with a phase offset of \( \pi/3 \) to asynchronous actuation for both mechanically coupled and uncoupled configurations. 
As shown in Fig.~\ref{fig:coupling}D, E, the mechanically coupled system rapidly converges towards the desired symmetric flapping (See Movie S3). 
In contrast, the uncoupled system maintains the initial phase offset throughout the duration of the experiment and is not able to generate symmetric wing flapping.

In addition, the mechanical coupling mechanism demonstrates strong robustness to asymmetries between the mechanical properties of the two wings (Fig.~S5). 
Similar to Fig.~\ref{fig:eigenvalues}F, we conducted experiments using a modified setup in which the passive wing was replaced by an inertial beam, allowing the effective inertia of the system to be adjusted easily and precisely. 
Using this configuration, we increased the inertia of one wing (\( I_1 \)) while keeping the inertia of the opposite wing (\( I_2 \)) unchanged, with the ratio \( I_1/I_2 \) ranging from 0.88 to 1.23. 
Consistent with earlier results, altering the inertia of one wing shifts its resonance frequency and, consequently, the emergent flapping frequency, which decreases from \(22.9~\mathrm{Hz}\) to \(20.2~\mathrm{Hz}\). 
However, when the wings are mechanically coupled, the system converges to a common wingbeat frequency that lies between the emergent frequencies of the left and right wings and maintains phase synchronization despite the inertia mismatch (Fig.~\ref{fig:coupling}G). 
Interestingly, the wing with the higher natural frequency exhibits a smaller flapping amplitude than its uncoupled emergent amplitude (Fig.~\ref{fig:coupling}H), while the wing with the lower natural frequency exhibits a larger amplitude. 
This amplitude redistribution provides a potential compensatory mechanism for mitigating differences in flapping frequency caused by inertia asymmetry.

In this coupled configuration, changes in inertia over a certain range lead to a global adjustment of both amplitude and frequency, while preserving left–right synchronization. 
This illustrates how mechanical coupling complements dSA: independent adaptation to mechanical changes at the wing level is retained, while synchronized coordination still emerges through the elastic coupling. 
Independent left–right actuation provides local adaptability and responsiveness, while mechanical coupling ensures coordinated, stable flapping. 
This architecture avoids centralized control, explicit phase commands, or position sensing, and instead leverages embodied dynamics to achieve coordination, mirroring the solution evolved by insects with asynchronous muscles \cite{Deora2021-hy}.

\subsection*{Flight robustness to wing damage and collisions} 

 \begin{figure}[htbp]
\centering
\includegraphics[scale=1]{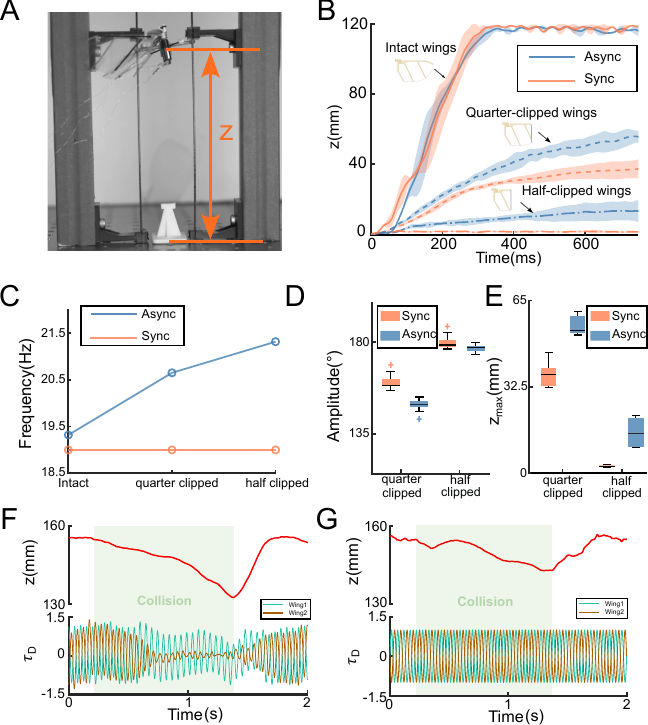}
\caption{Effects of wing collision and wing damage on asynchronous and synchronous actuation performance. \textbf{A} Photograph of the tethered takeoff platform. \textbf{B} Vertical position of the robot along the trackway as a function of time for intact and damaged wings. \textbf{C--E} Wingbeat frequency, wing stroke amplitude, and maximum height reached by the robot for intact and damaged wings, respectively. The orange lines and box plots correspond to synchronous actuation, while the blue lines and box plots correspond to asynchronous actuation. \textbf{F, G} Robot vertical position along the trackway and measured output force during a collision experiment. The green shaded region indicates the duration of contact. \textbf{F}: Asynchronous actuation. \textbf{G}: Synchronous actuation.}
\label{fig:ad&res}
\end{figure}

We previously characterized a fixed robot's response to wing collisions.
In this section, we evaluate the ability of dSA-actuated wings to respond to damage and collisions during vertical flight (Fig.~\ref{fig:ad&res}A). 
The robot is constrained to translate only along the vertical axis by two parallel carbon-fiber guide rods, while remaining free to generate lift through its own wingbeats. 
This setup allows investigation of flight dynamics during wing collisions and damage without needing to stabilize the hovering robot.

We first verify that the dSA-actuated robot is capable of sustained vertical flight in the absence of external perturbations. 
The robot first takes off on the vertical trackway with no obstacles present. 
A high-speed motion capture system (OptiTrack PrimeX~13) is used to record the vertical position \( z(t) \) throughout the experiment. 
Under both synchronous and asynchronous actuation, the robot successfully lifts off, ascends along the trackway at comparable rates, and reaches steady flight near the upper region of the trackway (\( z = 120~\mathrm{mm} \)) within \(0.3~\mathrm{s}\) (Fig.~\ref{fig:ad&res}B). 
These results confirm that the platform generates sufficient lift to sustain stable flight under nominal conditions.

We then artificially damaged both wings by removing either one-quarter or one-half of their length (see Movie S2). The vertical trackway experiment was repeated five times for each degree of wing damage under both asynchronous and synchronous actuation, without modifying the control parameters. Specifically, ($r_3$) and ($\mu$) were held constant for asynchronous actuation, while the flapping frequency and amplitude were kept unchanged for synchronous actuation. Motion capture measurements show that under asynchronous actuation (Fig.~\ref{fig:ad&res}C, D), the quarter-damaged wing exhibits a \(9\%\) higher flapping frequency and a \(5\%\) smaller flapping amplitude compared to the synchronous case during both takeoff and steady-state flight. 
This adaptive response enables the robot to reach a steady-state vertical position that is more than \(50\%\) higher for asynchronous flight with damaged wings as compared to the synchronous case (Fig.~\ref{fig:ad&res}B, E). 

Notably, under severe wing damage (half-cut wings), the synchronously actuated robot fails to generate sufficient lift to leave the bottom of the trackway. 
In contrast, the asynchronously actuated robot automatically operates at a \(12.6\%\) higher flapping frequency while maintaining approximately the same stroke amplitude, enabling ascent to a height of \(13 \pm 5~\mathrm{mm}\). 
This response closely resembles compensation strategies observed in flying insects \cite{rajabi2020insect, kassner2016kinematic, muijres2017flies}, many of which recover lost aerodynamic force through increased wingbeat frequency. 
For example, fruit flies increase their wingbeat frequency by approximately $9.2\%$ \cite{Salem2022-fg}, while hummingbird hawkmoths exhibit frequency increases of roughly $10.2\%$ \cite{kihlstrom2021wing} following a $25\%$ loss of wing area. 
These results suggest that the resonance-tracking behavior enabled by asynchronous actuation allows the system to partially compensate for wing loss without explicit damage detection, controller retuning, or feedback intervention.

In addition to compensating for gradual changes in wing mechanics, asynchronous actuation also provides a rapid, embodied control response to collisions. 
To test this capability under flight-like conditions, the robot is first allowed to ascend to the top of the vertical trackway.
An obstacle is then introduced into the path of the left wing.
In both synchronous and asynchronous cases the wing collides with the object and we observe a pronounced reduction in its flapping amplitude and velocity, while the right wing continues flapping (See Movie S2).
In the asynchronous case, the obstructed wing motion is significantly decreased relative to the synchronous actuation case (Fig.\ref{fig:ad&res}F, G). 
Under synchronous actuation, the wing continues to be driven at full driving force, resulting in large impact forces.

As the obstructed wing slows down more significantly under asynchronous actuation, the robot experiences a larger reduction in overall lift and vertical position. 
This greater loss of height likely assists disengagement from the obstacle by moving the wing away from the collision path. 
Once the obstacle is removed, the wing immediately resumes its nominal flapping behavior, allowing the robot to ascend back to the top of the trackway. 
This rapid and reversible response suggests that asynchronous actuation provides an inherent self-protection mechanism by localizing the effects of collisions and mitigating the transmission of reaction forces throughout the system.

Together, these vertical trackway experiments demonstrate that asynchronous actuation endows flapping-wing robots with intrinsic adaptability to both persistent mechanical changes (such as wing damage) and impulsive disturbances (such as collisions). 
While alternative approaches could achieve similar robustness using additional sensors and control logic, asynchronous actuation offers these benefits through passive dynamical interaction alone.

\subsection*{First free-flight of an asynchronous flapping wing robot} 
 \begin{figure}[htbp]
\centering
\includegraphics[width=1  \linewidth]{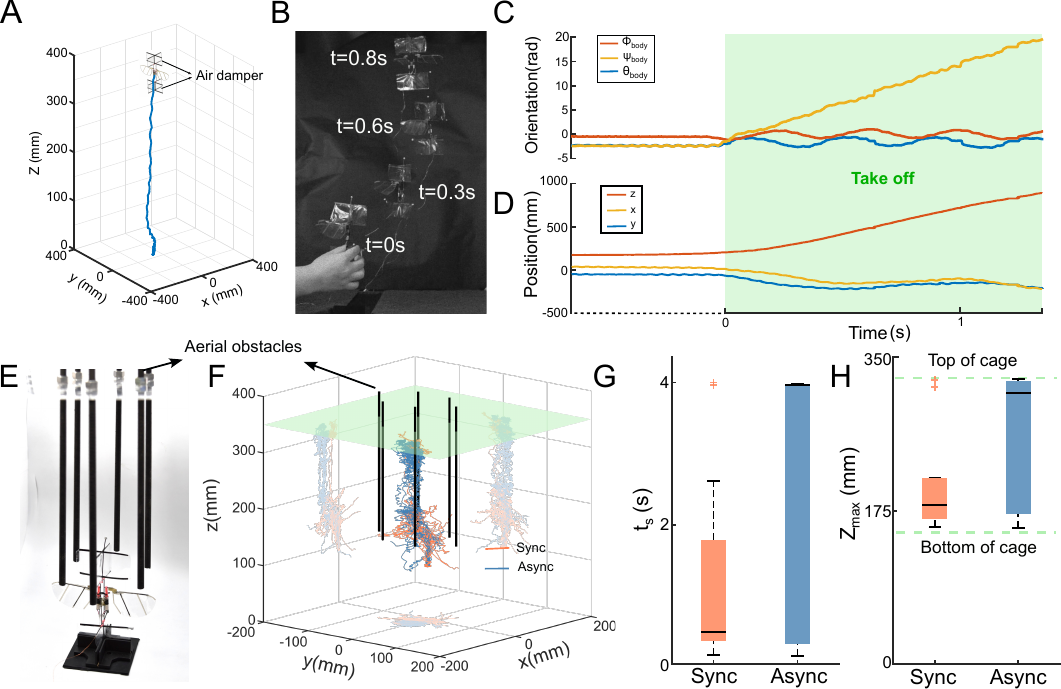}
\caption{Flight stability in complex environments. \textbf{A} Representative free-flight trajectory of the robot. \textbf{B} Sequential snapshots of the robot during a free takeoff experiment with the air damper attached. \textbf{C} Robot orientation in the world frame. \textbf{D} Robot position in the world frame. \textbf{E} Photograph of the cluttered flight environment used for the experiments. \textbf{F} Flight trajectories of robots operating with synchronous and asynchronous actuation in the cluttered environment. \textbf{G} Flight duration of the robot within the cage. \textbf{H} Maximum height reached by the robot within the cage.}
\label{fig:takeoff}
\end{figure}

To evaluate the practical feasibility of asynchronous actuation in aerial locomotion, we next demonstrate free flight of the robot and investigate the stability and resilience benefits of this actuation strategy under realistic operating conditions. 
As described previously, air dampers were incorporated to isolate the flight benefits provided by dSA without relying on any additional sensing, feedback, or active control strategies. 
With the dampers in place, the robot is capable of vertical take-off and hovering while maintaining passive attitude stability. 
Figure~\ref{fig:takeoff}A--D shows a representative free-flight experiment.

We first characterize flight performance in an open, unobstructed environment. 
Using a motion-capture system (OptiTrack PrimeX 13), we recorded the robot's position and orientation in the world frame during untethered flight (Movie~S4). 
A typical trajectory is presented in Fig.~\ref{fig:takeoff}A,D. Following take-off, the robot ascends approximately 50~cm within 1~s. 
Throughout the flight, body pitch and roll remain bounded with only small-amplitude oscillations (Fig.~\ref{fig:takeoff}C). 
The robot exhibits continuous yaw rotation, while the left and right wings maintain their nominal in-phase flapping pattern. 
These results demonstrate that asynchronous actuation, combined with passive aerodynamic stabilization, is sufficient to sustain stable untethered flight without centralized control. 
Moreover, it represents the first successful free-flight demonstration of a flapping-wing micro aerial vehicle driven by insect-inspired asynchronous actuation.

\subsection*{Asynchronous actuation improves flight performance in cluttered environments}

We next investigate whether asynchronous actuation improves flight robustness in complex environments involving frequent wing collisions. 
Our previous experiments demonstrated that dSA-actuated wings respond rapidly to collisions by reducing flapping amplitude and applied force. 
Based on this, we hypothesize that asynchronously actuated robots can perform better when near obstacles as compared to synchronously actuated robots.

To test this hypothesis, we constructed a cluttered flight environment consisting of a vertical hexagonal cage with rigid poles located at its vertices (Fig.~\ref{fig:takeoff}E). 
The face-to-face distance of the cage was 9.5~cm, corresponding to 60\% of the robot's wingspan. 
This geometry precludes collision-free hovering, as at least one wing must intermittently contact the poles regardless of the flight trajectory. 
The only feasible path to the top of the cage requires the robot to navigate its body between adjacent poles while the wings flap alternately inside and outside the enclosed region. 
The robot was released from the bottom of the cage, ensuring inevitable wing collisions during ascent and allowing us to evaluate flight stability in a cluttered environment (See Movie S4).

We performed 12 flight trials for each actuation strategy under identical experimental conditions. 
The resulting flight trajectories are summarized in Fig.~\ref{fig:takeoff}F--H. Under asynchronous actuation, 9 out of 12 trials successfully reached the top of the cage and remained airborne throughout the 4~s test duration. 
In contrast, only 2 out of 12 trials achieved the same outcome under synchronous actuation. 
These results demonstrate that asynchronous actuation substantially improves the robot's ability to maintain stable flight and traverse cluttered environments despite repeated wing collisions. 
Most failures under synchronous actuation occur within the first \(0.2~\mathrm{s}\) after the robot enters the cage, when initial wing collisions generate large reaction forces.
These forces induce substantial yaw, roll, and pitch perturbations, severely degrading flight stability (Fig.~S6). 
In several trials, repeated high-force impacts lead to extensive wing damage and complete flight termination. 
In contrast, asynchronous actuation localizes collision effects to the impacted wing and reduces sustained contact forces, thereby preserving overall flight stability.

Finally, we demonstrate the embodied responsiveness of asynchronous actuation to sudden external perturbations. 
As shown in Fig.~\ref{fig:intro} F, when an asynchronous robot is caught by hand during a free-flight, both wings immediately halt flapping, and resume flapping when released (See Movie S1). 
This behavior is predicted from our benchtop and tethered wing collision experiments. 
This catch and release experiment highlights an emergent safety mechanism enabled by asynchronous actuation: the capability for rapid, sensorless responses to large disturbances through intrinsic actuator dynamics.

\section{Discussion}

Flies, bees, and mosquitoes are some of the many insects in nature that use asynchronous muscles in which their wingbeats are generated autonomously without timing from the brain. 
In this work, we developed the first flapping wing robot capable of free-flight that uses asynchronous actuation.
This autonomous actuation method enabled the robot to adapt its wingbeat frequency to changes in body mechanics, and respond to collisions by reducing actuation torque and thus contact force. 
Asynchronous actuation enabled this responsiveness without requiring extra sensors or a centralized controller monitoring for collisions or changes in resonance.
This method of actuation substantially improved free-flight performance in a aerial clutter as compared to the traditional synchronous approach that all other flapping wing robots employ. 

Similar principles for adaptivity have been explored in other robotic systems, such as dynamic hopping robots that tune leg stiffness to match resonance \cite{bliss2012experimental}, or swimming robots that use central pattern generators (CPG) that adapt to body damage and environmental interactions \cite{li2014approach}. 
Overall, actuation strategies that utilize feedback-dependent oscillators rather than time-based commands exist in both insect \cite{bartussek2013limit} and robotic systems \cite{Bayiz2019-te} and can facilitate intrinsic responsiveness and robustness to perturbations. 

Damage mitigation is an important property for bioinspired flapping robots, as during prolonged operation, significant wing wear can accumulate due to repeated collisions and aerodynamic loading \cite{Foster2011-iw} leading to degraded flight performance and shortened operational lifetime \cite{Mountcastle2016-eb}. 
A variety of strategies have been developed to mitigate wing damage and enhance robustness. 
These include mechanical designs such as soft actuators \cite{Chen2021-ui, Chen2019-if}, foldable wings \cite{Mountcastle2019-un, Jankauski2022-bj}, and compliant transmission systems \cite{Gao2022-jb}, as well as sensing and control-based approaches for collision detection and avoidance \cite{park2023development}. 
In parallel, biological studies have investigated how insects maintain flight performance despite wing damage \cite{Salem2022-fg, Meng2023-jw, Le-Roy2019-uk, Fernandez2012-as, Foster2011-sg}, and robotic counterparts have explored compensatory control strategies \cite{Tu2021-sd}. 
The inherent safety that asynchronous actuation provides presents another mechanisms for wing damage mitigation, a preventative mechanism that is embodied in the actuator itself without the need to explicitly detect collisions. 
Our results show that asynchronous actuation reduces collision forces by rapidly suppressing actuation torque upon impact, effectively acting as an embodied protection mechanism. 
We hypothesize that a similar mechanism may exist in biological flight \cite{Burnett2023-cl}, where asynchronous muscle dynamics allow wing actuation to temporarily halt or attenuate during collisions, reducing structural damage.

Moreover, our method is able to compensate for wing damage and sustain relatively higher lift, allowing the robot to maintain flight through emergent adaptation of the wingbeat to the system’s resonant dynamics. 
Compared to the approaches above, asynchronous actuation offers a fundamentally different pathway: instead of explicitly detecting damage or reconfiguring control, the system passively adapts through its intrinsic dynamics. 
By embedding responsiveness directly into the actuator dynamics, insects—and by extension asynchronous actuated robots—may be able to react to perturbations faster than is possible through neural feedback alone.

In biological flyers, two separate sets of muscles drive the left and right wings, and mechanical synchronization is achieved through elastic coupling rather than centralized coordination \cite{Deora2015-ii, Deora2017-ht, Deora2021-hy}. 
Similar decentralized synchronization strategies have been observed in other robotic systems, such as coupled oscillators in legged locomotion and soft robots \cite{comoretto2025physical}, where coordination emerges from physical interactions rather than explicit communication \cite{zhou2021collective, hao2022proprioceptive}. 
Our results show that this approach allows each wing to independently respond to local disturbances while maintaining global coordination without centralized control involved \cite{chung2010neurobiologically}, resolving the apparent trade-off between adaptability and synchronization. 

Our results demonstrate that asynchronous actuation enables improved flight performance in cluttered aerial conditions by reducing collision-induced forces and amplitude upon impact automatically and allowing the robot to maintain stable flight despite repeated impacts. 
This behavior differs from conventional flapping-wing control strategies, where collisions often lead to large reaction forces, destabilization, and potential failure. 
Considering the minimal control requirements and the robot’s strong performance in cluttered environments, our results also suggest why some small insects, which employ extremely rapid wingbeats (some greater than 800 Hz), may rely on asynchronous wingbeat actuation as a rapid stabilization strategy.

Future work will explore integrating asynchronous actuation with higher-level sensing and control to enable both reactive and deliberate behaviors. 
Advances in soft actuators and artificial muscle technologies may further allow dSA-like dynamics to be embedded directly into the actuator material, eliminating the need for external computation. 
Scaling this approach to a more complex environment may also enable new forms of robust, adaptive behaviors. 
In conclusion, this work demonstrates that delayed stretch activation provides a powerful framework for flapping-wing actuation based on embodied intelligence. 
Rather than explicitly controlling motion through time-dependent commands, dSA shapes the dynamical interaction between actuator and structure, enabling adaptive, resilient behavior to emerge naturally. 
This perspective suggests a broader design paradigm for robotics, where robustness and adaptability are achieved not through increased sensing and computation, but through carefully engineered physical dynamics.

\section{Acknowledgements}

This work was supported by US National Science Foundation RAISE grant no. IOS-2100858 to S.S. and N.G.

\section{Materials and Methods}

\subsection{Implementation of delayed stretch activation through Simulink real-time}

The Simulink Real-Time module in MATLAB provides a convenient framework for implementing control algorithms through a DAQ system (National instruments PCIe-6343) with real-time execution. Motor drivers (Pololu TB9051FTG Single Brushed DC Motor Driver Carrier), which support drive and coast mode, are used to apply the desired PWM signals while enabling back-EMF-based velocity sensing. The dynamics of the asynchronous actuation are generalized by a second-order transfer function, allowing straightforward implementation in Simulink:
\begin{equation}
    G(s)=\frac{F(s)}{V(s)}=\frac{\mu\alpha_3}{s^2+\alpha_2 s+\alpha_3}
\end{equation}

where $\alpha_2=r_3(1+\kappa)$, $\alpha_3=\kappa r_3^2$, $F(s)$ and $V(s)$ denote the Laplace transforms of the actuation force and wing rotational velocity, respectively. As described in the main text, velocity feedback is obtained from the motor back-EMF. In practice, the voltage across the motor terminals is measured using analog input channels on the DAQ, capturing both the applied PWM signal and the induced back-EMF. Data acquisition is performed at a sampling rate of 10 kHz to ensure sufficient resolution.
Experimentally, the motor constant is identified as $K_e=0.0157~V/(rad/s)$, with a no-load speed of approximately 260 rad/s at 6 V, corresponding to a maximum back-EMF of about 4 V. Given a 15 V supply, the measured signal is filtered by truncating values above 8 V to isolate the back-EMF component. The truncated segments are then reconstructed via linear interpolation to obtain a continuous estimate of the back-EMF voltage, which is subsequently converted to rotational velocity.
The output of the transfer function (desired torque) is mapped to a PWM duty cycle, saturated, and transmitted to the motor driver via the DAQ at 2 kHz. In the experiments, the duty cycle is limited to 0.7, ensuring a 30\% low-level interval within each PWM period. This interval enables reliable observation of the back-EMF signal without interference from the drive voltage.

\subsection{Fabrication}

The flexure hinges are designed to sustain high-frequency flapping while enabling passive kinematic adaptation arising from the asynchronous actuation of the system. Each hinge is fabricated as a laminated composite consisting of five laser-cut layers: two fiberglass outer layers (0.03 mm each), a continuous PET film (5 mil) serving as the compliant joint, and two adhesive layers (Adhesives Inc. 7876) that bond the stack into a symmetric structure. This configuration localizes bending to predefined hinge regions while maintaining overall structural integrity. The wings are similarly fabricated using a three-layer laminate composed of a fiberglass layer (0.01 mm) that acts as the structural skeleton, a PET film layer, and an adhesive layer (Adhesives Inc. 7876). The wings, hinges, and 3D printed connectors are assembled via bonding to form the complete mechanism. A discrete spring element is glued to the connector and anchored to the chassis, providing parallel elasticity within the actuation pathway.


\subsection{Motion capture}
Motion capture of the flapping robot was performed using four OptiTrack PrimeX 13 cameras. The cameras were arranged to provide full coverage of the flapping workspace and calibrated for sub-millimeter spatial accuracy. Retroreflective markers were placed on key locations of the robot. Kinematic data were recorded at 400 Hz, providing sufficient resolution to capture the flapping dynamics. The recorded data were processed in Motive and Matlab to reconstruct 3D trajectories and extract angular motion. This setup enables precise quantification of the robot’s kinematic response, including transient behaviors such as collision response driven by asynchronous actuation.

\printbibliography

\end{document}